\documentclass[conference]{IEEEtran}
\IEEEoverridecommandlockouts
\usepackage[hidelinks]{hyperref}
\usepackage{amsfonts, amsmath, booktabs, graphicx, multirow, tabularx}

\title{\LARGE\bf ADAPTOOD: Uncertainty-Aware Fine-Tuning\\for Out-of-Distribution ECG Time Series Models}
\author{Sotirios Vavaroutas$^{1}$, Yu Yvonne Wu$^{2}$, Ali Etemad$^{3}$, Cecilia Mascolo$^{1}$%
\thanks{$^{1}$University of Cambridge, UK}
\thanks{$^{2}$Dartmouth College, USA}
\thanks{$^{3}$Queen’s University, Canada}
}

\begin{document}
\maketitle
\pagestyle{plain}
\thispagestyle{plain}
\begin{abstract}
Data samples used for training often differ from those encountered during fine-tuning and deployment, and while ML models show promise, their performance remains limited when only small annotated datasets are available. Performance often degrades under distribution shifts caused by diverse sensors, populations, and application settings. Although pre-training helps, models frequently encounter out-of-distribution (OOD) data in real-world settings, leading to reduced robustness. Existing adaptation methods usually assume fixed distribution shifts and struggle when multiple types or severities occur. In particular, they overlook shift \textit{severity}, for example treating adaptation to a large familiar dataset the same as adaptation to a small dataset with a new task, which limits generalisation. To address this, we propose ADAPTOOD, a novel framework that leverages \textit{data uncertainty} to quantify distribution shift severity and guide fine-tuning for time series. This uncertainty measures how strongly samples from the target deployment distribution deviate from the pre-training distribution, providing a direct signal of OOD severity. Our framework combines this uncertainty with low-rank model updates and adaptive hyperparameter optimisation to improve adaptation. We show that ADAPTOOD achieves up to 7\% higher accuracy and 12.9\% higher precision than existing methods in OOD tasks, maintaining strong performance as distribution shift severity increases.
\end{abstract}
\section{Introduction} \label{introduction}
With the increasing digitisation of real‑world systems, ML models trained on large‑scale sensor and signal data have shown strong potential for identifying complex patterns and irregularities~\cite{gupta24}. However, deploying such models outside controlled research settings remains difficult, largely due to the limited availability of high‑quality, labelled datasets required for robust generalisation~\cite{ding25}. Producing reliable annotations for time series data demands specialised expertise and is often time‑consuming~\cite{embc25}. This often leads to small datasets that might fail to capture the variability of real-world settings. While models trained from scratch on small datasets may perform well within a specific context, they tend to overfit and not generalise to new data due to the variety of settings in which data samples are collected.

Fine-tuning pre-trained models has shown promise in these cases, as it allows models to leverage general representations learned from large datasets before adapting to specific tasks with limited data, yet this strategy often struggles under out-of-distribution (OOD) conditions, where test data diverge from the training distribution due to population heterogeneity, differing sensor types, and variations in collection protocols~\cite{jingkang24, rajendran24}. These variations can result in significant distribution shifts, complicating the development of robust and generalisable models for biosignal interpretation~\cite{xia22}.

While transfer learning and domain adaptation methods have been proposed to mitigate such distributional differences~\cite{jun23}, they all face common challenges, including modality mismatch, limited labelled data, and inter-subject variability~\cite{roy19, lichao20}, which often lead to degraded performance and poor generalisation. These methods treat OOD cases under fixed assumptions without considering their granularity or \textit{severity}. They often apply the same adaptation regardless of shift levels or task complexity, so they often do not generalise to the broad spectrum of OOD data that may occur during fine-tuning~\cite{kouw19}, as most adopt a coarse-grained binary approach, treating data as either in-domain or out-of-domain. However, in practice, multiple sources of shifts with varying severities frequently overlap~\cite{jiahao24}, as for example a model trained on data from young adults might underperform on data from elderly patients recorded from a different device, exhibiting a compound population and sensor shift.

To address these challenges, we propose ADAPTOOD: an adaptation framework that quantifies distribution shift severity, and uses this information to guide model fine-tuning on OOD electrocardiograms (ECGs). To quantify severity, we leverage data uncertainty, which reflects how unfamiliar a target input is with regards to the pre-training distribution. The key intuition is that when a model encounters new OOD inputs that are dissimilar to its training distribution, its uncertainty level varies~\cite{lakshminarayanan17}. More specifically, a low uncertainty implies that the OOD dataset lies near the training distribution, indicating a mild shift, while a high uncertainty signals a more severe shift~\cite{ovadia19, svensson25}. We fine-tune the pre-trained model based on the guidance of this OOD severity quantification mechanism, so that the system adjusts how aggressively it learns from new input. To support this, ADAPTOOD also incorporates low-rank adaptation and adaptive hyperparameter optimisation, improving the fine-tuning both in effectiveness and in computational efficiency.

We evaluate ADAPTOOD using datasets that reflect realistic distribution shifts in ECG time series. We compare against transfer learning, supervised learning, and domain adaptation baselines, and against ablated versions of our approach. Across distribution shifts, ADAPTOOD consistently outperforms alternatives. Our key contributions are summarised below:

\begin{itemize}
    \item We introduce a novel mechanism that leverages data uncertainty to assess the severity of OOD data shifts in terms of their divergence from the pre-training data.
    \item We develop an OOD severity-flexible adaptation approach that uses uncertainty and hyperparameter tuning to achieve better calibration according to the severity of distribution shift, while also incorporating low-rank adaptation to maintain computational efficiency.
    \item Through comprehensive evaluation in OOD model adaptation, we demonstrate that our method achieves up to 7\% higher accuracy and 12.9\% higher precision compared to best-performing baselines, as well as efficiency-wise performance gains and consistently strong performance across metrics.
\end{itemize}
\section{Related Work} \label{related_work}
\textbf{Transfer Learning.}
Transfer learning pre-trains a model on one task and fine-tunes it on a related one~\cite{gholizade25}, leveraging learned representations to improve performance on smaller or less representative datasets~\cite{hosna22}. In time series analysis, this is particularly appealing due to the strong temporal dependencies shared across cardiac signals~\cite{ding25}, which can be learned during pre-training and reused across distribution-shifted signals from heterogeneous sources. Prior work demonstrates that models trained directly on small health datasets tend to generalise poorly~\cite{althnian21}, while model performance remains limited when faced with heterogeneous ECG scenarios~\cite{bizzego21}, where sensor types, patient populations and signal qualities may change simultaneously~\cite{li25}. Existing methods thus struggle to handle the diverse shift types common in time series data, constraining their generalisability~\cite{huang24}.

\textbf{Representation \& Contrastive Learning.}
Representation learning trains models to discover the most useful representations of input data by capturing robust embeddings~\cite{bengio12}. In time series, prior work has explored learning representations robust to latent and dynamically changing distributions~\cite{wang23} for OOD data. However, this expects the data to come from a limited set of known conditions, which is restrictive for ECG signals collected in the wild, where distribution shifts can be continuous and unpredictable~\cite{huang24}. Additionally, ECG-specific representation learning methods~\cite{mehari22} aim to learn task-agnostic embeddings transferable across patient populations and diagnostic tasks, but often rely on pretext tasks that may not fully capture temporal variability, noise, or morphology changes introduced by relevant sensors~\cite{trirat24}.

In parallel, recent contrastive approaches further incorporate clinical metadata to align ECG embeddings with clinically meaningful differences between subjects~\cite{shu25}. While these provide a strong and scalable starting point for downstream analysis, complementary mechanisms are required to support effective fine-tuning when encountering varying distribution shifts in cardiac data~\cite{li25}.

\textbf{Biosignal Domain Adaptation.}
Domain adaptation seeks to transfer knowledge from a source to a target domain~\cite{farahani21}. Prior work has explored domain adaptation without access to source data~\cite{chidlovskii16} and adaptation leveraging both temporal and frequency features~\cite{huan23}. Yet, in ECG applications, shifts arise from changes in acquisition hardware, sampling rates, or patient demographics often occurring concurrently, while such existing methods assume a fixed shift type and overlook its severity~\cite{jingkang24}.

\textbf{Hyperparameter Optimisation for Biosignal Model Fine-Tuning.}
Hyperparameter optimisation has been shown to improve generalisation across use cases~\cite{yang20}, with strong results under domain and subpopulation shifts using small OOD validation sets~\cite{choi24}. However, its application to fine-tuning models for biosignals and time series remains underexplored, which is particularly useful in ECG settings where the amount of available data and the nature of the downstream task can vary substantially~\cite{ding25}.

\textbf{Uncertainty Quantification for Distribution Shift.}
Uncertainty is imperative in model fine-tuning, as it often reflects deviations from the original training distribution. Prior work has shown the importance of uncertainty quantification~\cite{tianyu24}, its relevance in time series data~\cite{ziqi25}, and its behaviour under dataset shifts~\cite{ovadia19}. For ECGs, uncertainty can naturally capture fine-grained temporal and morphological variations induced by sensor noise, patient motion, or physiological differences~\cite{lopez25}, thus being helpful as a guiding signal for model adaptation or fine-tuning in time series settings.

Overall, existing methods address aspects of distribution shift but fail to account for its \text{severity} and heterogeneity, which are inherent to time series. In practice, adaptation must cope with limited labelled data and overlapping shift types, requiring both parameter-efficient updates and adaptive optimisation strategies. This motivates ADAPTOOD, which leverages uncertainty as a data-driven signal to quantify shift severity and guide fine-tuning, enabling robust adaptation in OOD scenarios.
\section{Methods} \label{methods}
\subsection{Problem Definition} \label{methods:problem_definition}
Distribution shifts challenge model deployment~\cite{huaxiu22}. These occur when the joint distribution of inputs and outputs, $P(x, y)$, encountered during testing ($D_t$), differs from that seen during training ($D_s$). Common sources include population shifts~\cite{yuzhe23}, where the marginal distribution $P(x)$ changes due to variation across diverse patient groups, or sensor shifts~\cite{simons20}, where data is recorded using different devices and leads to variations in $P(x|y)$. Additionally, label shifts arise due to changes in the task-specific label distribution $P(y)$, and temporal shifts due to a distribution drift over time, where $P(x, y)$ evolves due to changes in population health status or medical practice~\cite{spathis22}. Domain shifts are also important as they can occur when a model pre-trained on one signal, like ECG, is applied to another, such as photoplethysmogram (PPG)~\cite{rajendran24}. Finally, contextual shifts driven by changes in recording conditions (e.g., rest, exercise, surgery) alter physiological patterns like the heart rate and blood pressure~\cite{mouxiang24}, thus challenging generalisation.

\begin{figure*}[t]
    \centering
    \includegraphics[width = 0.8\linewidth]{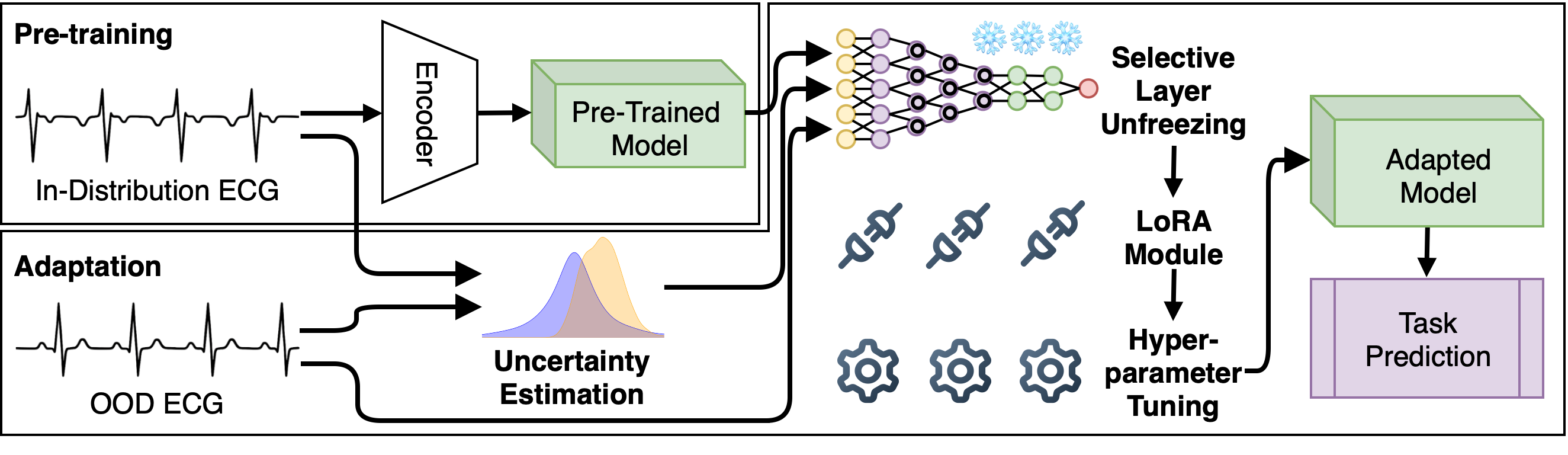}
    \caption{Overview of ADAPTOOD. The system adapts a pre-trained model to OOD data by first estimating uncertainty between source and target data, and then using this information to guide selective layer unfreezing. Subsequently, it applies LoRA for efficient fine-tuning and performs hyperparameter tuning for enhanced optimisation across downstream tasks.}
    \label{fig:system_overview}
\end{figure*}

\subsection{System Overview} \label{methods:system_overview}
ADAPTOOD adapts a pre-trained model to OOD data by first estimating uncertainty between source and target data, and then using this information to guide selective layer unfreezing. Subsequently, it applies low-rank adaptation (LoRA) for efficient fine-tuning and performs hyperparameter tuning for enhanced optimisation across downstream tasks, as depicted in Figure~\ref{fig:system_overview}. Using a classification model pre-trained on in-distribution data, it focuses on the fine-tuning stage to effectively and efficiently alleviate diverse OOD severity. Specifically, it includes an uncertainty module to quantify OOD severity levels, a Bayesian hyperparameter tuning module that adapts the model to optimise it to varying tasks, and a low-rank adaptation module (LoRA) to enable parameter-efficient performance. Using the estimated uncertainty to guide the smart unfreezing of model layers, combined with LoRA-based adaptation and Bayesian hyperparameter optimisation, it fine-tunes the model for robust generalisation on OOD data. In the following subsections, we describe each module.

\subsection{Uncertainty-Guided Model Adaptation} \label{methods:uncertainty}
Uncertainty estimation is fundamental to ADAPTOOD. In our framework, uncertainty reflects the divergence between the pre-training data distribution $D_s$ and the targeted OOD data $D_t$, by measuring the degree of overlap between them. Specifically, we capture $\sigma(D_t \mid P(D_s))$, representing the uncertainty of the target data with respect to the pre-training source distribution. This measures the degree of overlap (or divergence) between the two distributions, thus serving as a proxy for detecting distributional shifts~\cite{svensson25}. A high uncertainty can indicate a severe OOD shift, while a low uncertainty suggests greater similarity to the source distribution~\cite{ovadia19}.

To quantify the shift between $D_s$ and $D_t$, we employ data uncertainty estimation. We apply Principal Component Analysis (PCA) to reduce high-dimensional ECG signals into a single dominant component, to improve the effective density estimation while retaining the key variance in the data. We then estimate the uncertainty $\sigma(D_t \mid P(D_s))$ by calculating distribution-level divergence. This is done by leveraging the Mahalanobis ($d_m(x)$) and Hellinger ($d_h(x)$) distances as estimation metrics, discussed below. We apply Gaussian Kernel Density Estimation (KDE) to both the training and OOD dataset to estimate their respective probability density functions, and compute the uncertainty:
\begin{equation}
\sigma(D_t \mid P(D_s)) \gets \{d_h(D_t, D_s), d_m(D_t, D_s)\}
\end{equation}

\textbf{Mahalanobis distances} are a common way of measuring the distance between distributions, considering their correlation~\cite{maesschalck20}. To calculate, we compute from the pre-training data the mean vector $y$ and the covariance matrix $\Sigma$ that represent the centre and spread of the data in the feature space. Using these, we calculate the Mahalanobis distance to measure how far each new batch is from the centre of the known distribution~\cite{venkataramanan23}. As such, we use this metric to efficiently quantify OOD severity, where a larger value indicates a higher level of uncertainty, as it reflects the data's deviation from the known distribution and, thus, the OOD severity.

\textbf{Hellinger distances} similarly calculate the similarity between two probability distributions~\cite{govindaraj22}. They have been previously used to quantify uncertainty in stochastic differential equations with non-Gaussian parameters by minimising the distance between empirical probability densities~\cite{zheng21}. In our case, they can quantify how much an OOD distribution diverges from the training distribution. Thus, the greater the distance, the more severe the OOD shift. For each incoming batch of samples, we compute the Hellinger distance by comparing the probability density of the new samples against the source.

\textbf{Choice of Complementary Distance-Based Metrics.} We use these two metrics for capturing complementary aspects of uncertainty related to OOD severity. For an ablation study comparing them, refer to Section~\ref{results:ablation_study}. Mahalanobis distances quantify how far OOD batches deviate from the known distribution, incorporating feature correlations and scale~\cite{maesschalck20}. Hellinger distances, by contrast, measure divergence between overall probability distributions, capturing global distributional shifts~\cite{govindaraj22}. Using both metrics gives a more complete picture of OOD severity than a single metric, and we chose them as they account for correlations between variables, unlike simpler metrics like the Euclidean distance, which only focuses on straight-line distances between two points in feature space and does not account for the scale of features~\cite{danielsson80}. Additionally, unlike alternatives like MC-dropout, which estimates uncertainty by running multiple stochastic forward passes with dropout active~\cite{gal16}, these metrics calculate uncertainty before fine-tuning. Thus, they operate pre-inference, allowing for proactive severity estimation.

For computing our final uncertainty metric, given $d_h(D_t, D_s)$ and $d_m(D_t, D_s)$ along with their respective maximum values $d_h^{\max}$ and $d_m^{\max}$, we calculate the normalised distances as follows:
\begin{equation}
    \tilde{d}_h = \frac{d_h(D_t, D_s)}{d_h^{\max}}, \quad
    \tilde{d}_m = \frac{d_m(D_t, D_s)}{d_m^{\max}}
\end{equation}

\noindent The combined weighted uncertainty metric is then:
\begin{equation}
    \sigma(D_t \mid P(D_s)) = D_{\text{combined}} = w \cdot \tilde{d}_h + (1 - w) \cdot \tilde{d}_m
\end{equation}

Here, $w \in [0, 1]$ is the weight that controls the relative importance assigned to each distance. In our experiments, we set $w = 0.5$ to equally balance the two measures, reflecting a neutral stance when no prior knowledge favours one form of distributional discrepancy over the other. However, this can be adjusted to emphasise either distance metric depending on the characteristics of the source and target domains. Similarly, we standardise both $d_h^{\max}$ and $d_m^{\max}$ to 10, as in the range of datasets we evaluated these values consistently captured the upper bounds of uncertainty without saturation. These aspects remain configurable, useful if applied to datasets with substantially different statistical or uncertainty properties. The combined uncertainty score $D_{\text{combined}}$ is then used to inform the degree of selective fine-tuning as described in the next section.

\subsection{Selective Layer Unfreezing} \label{methods:layers}
Following the OOD severity estimation, as described above, ADAPTOOD uses this information to enable effective representation and parameter transfer from the pre-trained to the fine-tuning encoder. Existing methods typically fine-tune either the entire model or a fixed subset without considering how severe the OOD shift is, leading to unnecessary computational cost and reduced performance due to overfitting when the severity is low or underfitting when it is high. To address this, ADAPTOOD initially freezes all layers of the loaded model and then unfreezes a subset based on the OOD severity. This ensures adaptation without the risks of full retraining, such as excessive cost or catastrophic forgetting of in-distribution knowledge. To determine how many layers to unfreeze, ADAPTOOD uses a combination of linear and exponential strategies, and then unfreezes that number of layers starting from the upper layers closest to the output.

\textbf{Linear Calculation.}
In the linear approach, the number of layers to unfreeze is proportional to the normalised uncertainty between the source and target domains. Given a total number of layers $L$, this is done as follows:
\begin{equation}
    L_{\text{linear}} = (1 - D_{\text{combined}}) \cdot L
\end{equation}
\textbf{Exponential Calculation.} To allow for more sensitivity to uncertainty, an exponential decay function is applied using a decay factor $\alpha > 0$, which governs how rapidly the unfreezing rate drops with increasing uncertainty. The number of layers to unfreeze is:
\begin{equation}
    L_{\text{exponential}} = L \cdot e^{-\alpha \cdot D_{\text{combined}}}
\end{equation}
\textbf{Final Calculation.} The final number of layers that are made trainable by ADAPTOOD is taken as the mathematical mean between the linear and the exponential approach:
\begin{equation}
    L_{\text{final}} = \frac{L_{\text{linear}} + L_{\text{exponential}}}{2}
\end{equation}

Using this strategy, ADAPTOOD customises fine-tuning based on OOD severity. Of note, once the trainable layers are selected, the final layers are modified to match the target output shape, ensuring architectural compatibility with the new inputs and outputs.

\subsection{Low-Rank Adaptation} \label{methods:lora}
For effectiveness, we add Low-Rank Adaptation (LoRA) support to all the Conv1D layers found in the model, as these constitute the majority of the architecture's core processing units for capturing local temporal patterns. This is important in our time series tasks: since Conv1D layers are responsible for learning feature representations across time steps~\cite{kiranyaz21}, adapting them directly allows focusing the model's capacity where it matters the most, while keeping the rest of the architecture intact.

With LoRA, we also improve parameter efficiency. Instead of fine-tuning all parameters, LoRA freezes the pre-trained weights and introduces lightweight trainable components in the form of low-rank matrices~\cite{hu22}. Across all datasets, LoRA consistently leads to smaller model sizes in terms of memory footprint, as per Section~\ref{results:efficiency}, which is beneficial for deployment and reduces the risk of overfitting when working with limited target data.

\subsection{Hyperparameter Optimisation} \label{methods:hp_tuning}
We further optimise the model through dynamic hyperparameter tuning making it even more tailored to the target task with no need for manual adjustments. This is necessary as the hyperparameters performing well during pre-training are not always optimal under OOD. To this end, we use Bayesian optimisation~\cite{snoek12}, a widely-used approach that iteratively leverages information from prior trials to guide subsequent exploration~\cite{roy23}. Unlike less guided strategies such as random search, which focus on multiple local optima, this targets promising regions of the search space, yielding more consistent performance~\cite{snoek12}. This is valuable in OOD settings as it enables efficient identification of hyperparameter configurations that best accommodate distribution shifts.

We define a search space over the key dimensions of the optimiser and the learning rate. The optimiser is selected from a pool including Adam, AdaDelta, AdaGrad, AdaMax, and RMSprop, representing a diverse spectrum. Each has differing behaviours on sparse gradients and noise sensitivity, which can affect OOD adaptation performanc. For learning rate selection, we sample from a continuous range using logarithmic sampling to ensure finer-grained search. ADAPTOOD explores the search space for 10 trials, representing a pragmatic trade-off between thoroughness and efficiency. In our experiments, this was sufficient to observe meaningful improvements while keeping resource use reasonable. Once the search concludes, we retrieve the best configuration based on validation accuracy.

\subsection{Fine-Tuning} \label{methods:fine_tuning}
After incorporating the above adaptation mechanisms into ADAPTOOD, we proceed to the final stage: fine-tuning the model on the small labelled OOD dataset. During this phase, only the layers deemed trainable by the uncertainty-aware selection process are updated, ensuring focused and efficient adaptation. The model is trained for 20 epochs to balance performance gains with the risk of overfitting. This completes ADAPTOOD's model transfer process.

\begin{table*}[t]
    \centering
    \caption{Distribution shifts across datasets in comparison to the pre-training dataset}
    \label{tab:distribution_shifts}
    \footnotesize
    \begin{tabular}{>{\arraybackslash}m{3.5cm}|@{}
                    >{\centering\arraybackslash}m{2cm}@{}
                    >{\centering\arraybackslash}m{2.5cm}@{}
                    >{\centering\arraybackslash}m{2cm}@{}
                    >{\centering\arraybackslash}m{2cm}@{}
                    >{\centering\arraybackslash}m{2cm}@{}
                    >{\centering\arraybackslash}m{3cm}@{}}
    \toprule\toprule
        & \textbf{Sensor Shift} & \textbf{Population Shift} & \textbf{Temporal Shift} & \textbf{Label Shift} & \textbf{Modality Shift} & \textbf{Dimensionality Shift} \\
        \midrule\midrule
        ECG MIT-BIH & $\checkmark$ & $\checkmark$ & $\checkmark$ & & & \\
        ECG PTB-DB & $\checkmark$ & $\checkmark$ & $\checkmark$ & & & \\
        ECG MIMICPERformTT & $\checkmark$ & $\checkmark$ & $\checkmark$ & $\checkmark$ & & \\
        PPG MIMICPERformTT & $\checkmark$ & $\checkmark$ & $\checkmark$ & $\checkmark$ & $\checkmark$ & \\
        ECG CODEtest & $\checkmark$ & $\checkmark$ & $\checkmark$ & $\checkmark$ & & $\checkmark$ \\
        \bottomrule\bottomrule
    \end{tabular}
\end{table*}
\section{Experimental Setup} \label{experimental_setup}
\subsection{Model Architecture} \label{experimental_setup:model_architecture}
We adopt a 1D convolutional neural network (CNN) as our pre-trained model architecture due to its strong inductive bias for temporal signal processing, computational efficiency, and widespread use in time series analysis, where it can capture local patterns while remaining scalable to long sequences. The architecture starts with an input layer accepting vectors in the shape of the data, followed by several stacked Conv1D layers with increasing filters (32, 64, 128, 256, 512) and a kernel size of 5, each using ReLU activation and the same padding. Subsequently, MaxPooling1D layers reduce the spatial dimension after each convolution, and Dropout layers are inserted after higher-capacity layers to prevent overfitting. After the final convolution, the output is flattened and passed through two fully connected dense layers with 64 and 32 neurons respectively, both using the ReLU activation function, followed by a final dense output layer with a single neuron and a sigmoid activation function for classification. The model is then compiled using the settings chosen dynamically by the hyperaparameter tuner of Section~\ref{methods:hp_tuning}.

\subsection{Datasets} \label{experimental_setup:datasets}
We focus on ECG datasets, as they provide structured temporal signals whose dynamic and subtle variations make them ideal for evaluating model adaptation. To keep the task realistic for effective adaptation, we fine-tune on only a small subset of each dataset. We pre-train a CNN model on the PhysioNet Computing in Cardiology (CinC) 2017 dataset~\cite{clifford17}, with the version used~\cite{clifford17dataset} providing a clinically relevant task of classifying atrial fibrillation versus non-atrial fibrillation. This was selected for pre-training due to its well-defined diagnostic labels and task specificity, which provide a strong inductive prior for downstream cardiac classification tasks. We then evaluate adaptation across multiple downstream datasets introducing population, sensor, temporal, modality, label, and dimensionality shifts, including the PhysioNet MIT-BIH Arrhythmia Database~\cite{moody01}, the PhysioNet PTB-DB Database~\cite{bousseljot95}, the MIMICPERform ECG and PPG datasets~\cite{moody20,charlton22}, and the CODEtest Dataset~\cite{ribeiro20}. Across these datasets, tasks include distinguishing healthy from non-healthy samples and adult from neonate samples, while simulating data-limited and low-resource settings using small fine-tuning subsets. The exact type of distribution shift present in each case relative to the pre-training data is noted below, and summarised in Table~\ref{tab:distribution_shifts}.

\textbf{PhysioNet Computing in Cardiology (CinC) 2017.}
This dataset contains 8,528 single-lead ECG recordings~\cite{clifford17} and the version used~\cite{clifford17dataset} is sampled at 125Hz. We use it to pre-train a CNN model, which serves as the foundation for subsequent experiments. The pre-training task involves classifying the signals into atrial fibrillation and non-atrial fibrillation. As a clinically relevant dataset, this was selected for pre-training due to its well-defined diagnostic labels and task specificity, which provide a strong inductive prior for downstream cardiac classification tasks.

\textbf{PhysioNet MIT-BIH Arrhythmia Database.}
This is derived from MIT-BIH's arrhythmia database~\cite{moody01, moody01dataset}, and contains 109\,446 ECGs sampled at 125 Hz and segmented into 188 time steps. We use a single-lead version~\cite{fazeli18dataset} to better focus on features relevant to real-world wearable sensors, and randomly select 1,000 samples for fine-tuning to simulate a data-limited setting. This tests ADAPTOOD under constrained data, with our task distinguishing healthy from non-healthy cases. For evaluation, we use the full test set of 21,892 samples. Compared to the pre-training data, this introduces sensor shifts due to different wearable acquisition devices, population shifts due to different patient cohorts, and temporal shifts from changes in sampling contexts.

\textbf{PhysioNet PTB-DB Database.}
This dataset is extracted from the PTB diagnostic database~\cite{bousseljot95, bousseljot95dataset} and the version used~\cite{fazeli18dataset} includes 14,552 single-lead ECGs, sampled at 125Hz. We retain only 1,000 randomly-selected samples so that the task is realistic, distinguishing healthy from non-healthy samples. This introduces population, sensor, and temporal shifts in signal dynamics.

\textbf{MIMICPERform ECG Dataset.}
This dataset is extracted from the MIMIC-III waveform database~\cite{moody20}, and consists the MIMICPERform training and the MIMICPERform testing set~\cite{charlton22, charlton22dataset}. The task is to distinguish adult from neonate samples, and  we further limit the training set to 100 samples to showcase low-resource settings where large-scale training is not feasible. This dataset presents population (age and condition differences), label, sensor, and temporal (clinical context) shifts.

\textbf{MIMICPERform PPG Dataset.}
This contains PPG recordings from the same patients and timeframes as the MIMICPERform ECG set described above. While it does not include ECG signals,  we use it to test ADAPTOOD's ability to adapt across mobile and wearable biosignal modalities. This introduces a modality shift, since the ECG pre-trained model is now applied to PPG data, which have different physiological properties as they measure blood volume changes at the skin surface using light rather than electrical cardiac activity. This is the most challenging shift, and although ECG models are not expected to perform well on such fundamentally different signals, this setup provides insights into the limits of generalisation across biosignals.

\textbf{CODEtest Dataset.}
This dataset is the test version of the Clinical Outcomes in Digital Electrocardiology (CODE) study~\cite{ribeiro20, ribeiro20dataset} and includes 827 ECG recordings, each consisting of 12 leads sampled at 400Hz, with 4,096 data points per lead. We classify the samples into healthy and non-healthy groups. The recordings in this dataset have been annotated by cardiologists, medical students and others, and we use the gold standard annotation provided in the original source~\cite{ribeiro20dataset}. Further to population, sensor, and temporal shifts, this dataset also introduces a dimensionality shift by increasing the number of input channels. This is in contrast to previous cases, making it valuable for assessing ADAPTOOD.

\begin{table*}[t]
    \centering
    \caption{Evaluation results from OOD classification tasks on baselines and our ADAPTOOD proposed method}
    \label{tab:results}
    \begin{tabular}{l|cccccc}
        \toprule\toprule
        \textbf{Method} & \textbf{Accuracy} & \textbf{Precision} & \textbf{Recall} & \textbf{Sensitivity} & \textbf{Specificity} & \textbf{F1 Score}\\
        \midrule\midrule
        {} & \multicolumn{6}{c}{ECG MIT-BIH}\\
        \midrule
            Transfer Learning & 0.910\,\scriptsize{$\pm$\,0.007} & 0.891\,\scriptsize{$\pm$\,0.045} & 0.786\,\scriptsize{$\pm$\,0.055} & 0.595\,\scriptsize{$\pm$\,0.132} & 0.976\,\scriptsize{$\pm$\,0.024} & 0.818\,\scriptsize{$\pm$\,0.030}\\
            Supervised Learning & 0.906\,\scriptsize{$\pm$\,0.019} & 0.914\,\scriptsize{$\pm$\,0.009} & 0.745\,\scriptsize{$\pm$\,0.056} & 0.499\,\scriptsize{$\pm$\,0.114} & \textbf{0.991\,\scriptsize{$\pm$\,0.004}} & 0.793\,\scriptsize{$\pm$\,0.056}\\
            Feature-Based Domain Adaptation & 0.940\,\scriptsize{$\pm$\,0.002} & 0.921\,\scriptsize{$\pm$\,0.003} & 0.861\,\scriptsize{$\pm$\,0.011} & 0.740\,\scriptsize{$\pm$\,0.023} & 0.982\,\scriptsize{$\pm$\,0.003} & 0.887\,\scriptsize{$\pm$\,0.006}\\
            Instance-Based Domain Adaptation & 0.936\,\scriptsize{$\pm$\,0.004} & 0.921\,\scriptsize{$\pm$\,0.003} & 0.847\,\scriptsize{$\pm$\,0.019} & 0.710\,\scriptsize{$\pm$\,0.039} & 0.983\,\scriptsize{$\pm$\,0.003} & 0.878\,\scriptsize{$\pm$\,0.012}\\
            ADAPTOOD & \textbf{0.953\,\scriptsize{$\pm$\,0.002}} & \textbf{0.930\,\scriptsize{$\pm$\,0.008}} & \textbf{0.900\,\scriptsize{$\pm$\,0.005}} & \textbf{0.820\,\scriptsize{$\pm$\,0.013}} & 0.980\,\scriptsize{$\pm$\,0.005} & \textbf{0.914\,\scriptsize{$\pm$\,0.003}}\\
        \midrule\midrule
        {} & \multicolumn{6}{c}{ECG PTB-DB}\\
        \midrule
            Transfer Learning & 0.803\,\scriptsize{$\pm$\,0.033} & 0.772\,\scriptsize{$\pm$\,0.030} & 0.797\,\scriptsize{$\pm$\,0.023} & 0.811\,\scriptsize{$\pm$\,0.096} & 0.784\,\scriptsize{$\pm$\,0.139} & 0.769\,\scriptsize{$\pm$\,0.015}\\
            Supervised Learning & 0.818\,\scriptsize{$\pm$\,0.007} & 0.786\,\scriptsize{$\pm$\,0.007} & 0.728\,\scriptsize{$\pm$\,0.037} & 0.924\,\scriptsize{$\pm$\,0.028} & 0.531\,\scriptsize{$\pm$\,0.102} & 0.745\,\scriptsize{$\pm$\,0.028}\\
            Feature-Based Domain Adaptation & 0.875\,\scriptsize{$\pm$\,0.000} & 0.851\,\scriptsize{$\pm$\,0.013} & 0.836\,\scriptsize{$\pm$\,0.038} & 0.920\,\scriptsize{$\pm$\,0.045} & 0.753\,\scriptsize{$\pm$\,0.120} & 0.838\,\scriptsize{$\pm$\,0.012}\\
            Instance-Based Domain Adaptation & 0.873\,\scriptsize{$\pm$\,0.003} & 0.843\,\scriptsize{$\pm$\,0.007} & 0.835\,\scriptsize{$\pm$\,0.025} & 0.918\,\scriptsize{$\pm$\,0.024} & 0.753\,\scriptsize{$\pm$\,0.074} & 0.837\,\scriptsize{$\pm$\,0.011}\\
            ADAPTOOD & \textbf{0.922\,\scriptsize{$\pm$\,0.013}} & \textbf{0.901\,\scriptsize{$\pm$\,0.019}} & \textbf{0.899\,\scriptsize{$\pm$\,0.015}} & \textbf{0.948\,\scriptsize{$\pm$\,0.014}} & \textbf{0.852\,\scriptsize{$\pm$\,0.019}} & \textbf{0.900\,\scriptsize{$\pm$\,0.016}}\\
        \midrule\midrule
        {} & \multicolumn{6}{c}{ECG MIMICPERformTT}\\
        \midrule
            Transfer Learning & 0.882\,\scriptsize{$\pm$\,0.025} & 0.895\,\scriptsize{$\pm$\,0.015} & 0.882\,\scriptsize{$\pm$\,0.025} & 0.950\,\scriptsize{$\pm$\,0.045} & 0.813\,\scriptsize{$\pm$\,0.095} & 0.881\,\scriptsize{$\pm$\,0.026}\\
            Supervised Learning & 0.862\,\scriptsize{$\pm$\,0.033} & 0.874\,\scriptsize{$\pm$\,0.035} & 0.862\,\scriptsize{$\pm$\,0.033} & 0.920\,\scriptsize{$\pm$\,0.090} & 0.803\,\scriptsize{$\pm$\,0.075} & 0.861\,\scriptsize{$\pm$\,0.033}\\
            Feature-Based Domain Adaptation & 0.867\,\scriptsize{$\pm$\,0.003} & 0.869\,\scriptsize{$\pm$\,0.003} & 0.867\,\scriptsize{$\pm$\,0.003} & 0.827\,\scriptsize{$\pm$\,0.020} & \textbf{0.907\,\scriptsize{$\pm$\,0.020}} & 0.866\,\scriptsize{$\pm$\,0.003}\\
            Instance-Based Domain Adaptation & 0.842\,\scriptsize{$\pm$\,0.008} & 0.851\,\scriptsize{$\pm$\,0.008} & 0.842\,\scriptsize{$\pm$\,0.008} & 0.923\,\scriptsize{$\pm$\,0.010} & 0.760\,\scriptsize{$\pm$\,0.010} & 0.841\,\scriptsize{$\pm$\,0.008}\\
            ADAPTOOD & \textbf{0.942\,\scriptsize{$\pm$\,0.003}} & \textbf{0.944\,\scriptsize{$\pm$\,0.002}} & \textbf{0.942\,\scriptsize{$\pm$\,0.003}} & \textbf{0.980\,\scriptsize{$\pm$\,0.010}} & 0.903\,\scriptsize{$\pm$\,0.015} & \textbf{0.942\,\scriptsize{$\pm$\,0.003}}\\
        \midrule\midrule
        {} & \multicolumn{6}{c}{PPG MIMICPERformTT}\\
        \midrule
            Transfer Learning & 0.917\,\scriptsize{$\pm$\,0.008} & 0.929\,\scriptsize{$\pm$\,0.006} & 0.917\,\scriptsize{$\pm$\,0.008} & \textbf{1.000\,\scriptsize{$\pm$\,0.000}} & 0.833\,\scriptsize{$\pm$\,0.015} & 0.916\,\scriptsize{$\pm$\,0.008}\\
            Supervised Learning & 0.905\,\scriptsize{$\pm$\,0.015} & 0.920\,\scriptsize{$\pm$\,0.011} & 0.905\,\scriptsize{$\pm$\,0.015} & \textbf{1.000\,\scriptsize{$\pm$\,0.000}} & 0.810\,\scriptsize{$\pm$\,0.030} & 0.904\,\scriptsize{$\pm$\,0.016}\\
            Feature-Based Domain Adaptation & 0.942\,\scriptsize{$\pm$\,0.003} & 0.946\,\scriptsize{$\pm$\,0.004} & 0.942\,\scriptsize{$\pm$\,0.003} & 0.990\,\scriptsize{$\pm$\,0.010} & 0.893\,\scriptsize{$\pm$\,0.005} & 0.942\,\scriptsize{$\pm$\,0.003}\\
            Instance-Based Domain Adaptation & 0.945\,\scriptsize{$\pm$\,0.005} & 0.948\,\scriptsize{$\pm$\,0.007} & 0.945\,\scriptsize{$\pm$\,0.005} & 0.983\,\scriptsize{$\pm$\,0.015} & 0.907\,\scriptsize{$\pm$\,0.005} & 0.945\,\scriptsize{$\pm$\,0.005}\\
            ADAPTOOD & \textbf{0.975\,\scriptsize{$\pm$\,0.010}} & \textbf{0.976\,\scriptsize{$\pm$\,0.009}} & \textbf{0.975\,\scriptsize{$\pm$\,0.010}} & 0.995\,\scriptsize{$\pm$\,0.005} & \textbf{0.953\,\scriptsize{$\pm$\,0.025}} & \textbf{0.975\,\scriptsize{$\pm$\,0.010}}\\
        \midrule\midrule
        {} & \multicolumn{6}{c}{ECG CODEtest}\\
        \midrule
            Transfer Learning & 0.873\,\scriptsize{$\pm$\,0.007} & 0.825\,\scriptsize{$\pm$\,0.057} & 0.757\,\scriptsize{$\pm$\,0.050} & 0.565\,\scriptsize{$\pm$\,0.143} & 0.949\,\scriptsize{$\pm$\,0.043} & 0.773\,\scriptsize{$\pm$\,0.013}\\
            Supervised Learning & 0.861\,\scriptsize{$\pm$\,0.019} & 0.825\,\scriptsize{$\pm$\,0.027} & 0.669\,\scriptsize{$\pm$\,0.034} & 0.361\,\scriptsize{$\pm$\,0.069} & \textbf{0.978\,\scriptsize{$\pm$\,0.001}} & 0.706\,\scriptsize{$\pm$\,0.035}\\
            Feature-Based Domain Adaptation & 0.894\,\scriptsize{$\pm$\,0.006} & 0.803\,\scriptsize{$\pm$\,0.007} & 0.778\,\scriptsize{$\pm$\,0.025} & 0.611\,\scriptsize{$\pm$\,0.063} & 0.945\,\scriptsize{$\pm$\,0.012} & 0.789\,\scriptsize{$\pm$\,0.016}\\
            Instance-Based Domain Adaptation & 0.882\,\scriptsize{$\pm$\,0.046} & 0.765\,\scriptsize{$\pm$\,0.082} & 0.734\,\scriptsize{$\pm$\,0.043} & 0.528\,\scriptsize{$\pm$\,0.054} & 0.939\,\scriptsize{$\pm$\,0.033} & 0.746\,\scriptsize{$\pm$\,0.058}\\
            ADAPTOOD & \textbf{0.922\,\scriptsize{$\pm$\,0.018}} & \textbf{0.875\,\scriptsize{$\pm$\,0.053}} & \textbf{0.807\,\scriptsize{$\pm$\,0.052}} & \textbf{0.641\,\scriptsize{$\pm$\,0.093}} & 0.974\,\scriptsize{$\pm$\,0.011} & \textbf{0.836\,\scriptsize{$\pm$\,0.053}}\\
        \bottomrule\bottomrule
    \end{tabular}
\end{table*}

\subsection{Baselines \& Ablations} \label{experimental_setup:baselines}
To evaluate our approach, we establish key alternatives spanning conventional transfer learning, supervised learning, domain adaptation, and ablation settings. We compare against a transfer learning setup with frozen feature extractors, a supervised learning baseline using the same deep convolutional architecture as ADAPTOOD, a feature-based domain adaptation using the widely-used PRED strategy~\cite{daume07}, and an instance-based domain adaptation via nearest neighbours weighting~\cite{loog21}. In addition, we conduct ablation studies by using only the Hellinger distance, by using only the Mahalanobis distance, by removing hyperparameter tuning, and by disabling LoRA, in order to isolate the contribution of each component of ADAPTOOD. Implementation details for all baselines and ablations are provided below.

\textbf{Transfer Learning.}
This baseline reflects a conventional transfer learning setup and operates by fine-tuning a pre-trained model to extract features from the input data. A subset of the model's earlier layers serve as a feature extractor, with their weights frozen to preserve learned representations. The remaining layers are trainable, and the model is compiled using the original optimiser and loss function.

\textbf{Supervised Learning.}
This baseline works in a supervised fashion using a deep convolutional neural network with the same model layers as ADAPTOOD, described in Section~\ref{experimental_setup:model_architecture}. It is compiled using a cross-entropy loss and the Adam optimiser~\cite{kingma14}. Across all experiments, this baseline has access to the same total amount of samples as the alternatives, representing a fair comparison of how the model would perform when there is a only a niche dataset annotated.

\textbf{Feature-Based Domain Adaptation.}
This baseline illustrates a feature-based domain adaptation approach, which builds a shared feature representation to correct the difference between the source and target distributions. The task is then learned in this encoded space. For this baseline we use the ADAPT toolbox~\cite{mathelin25}, and specifically its PRED strategy~\cite{daume07} that is widely-used and frequently cited. This first trains a model on the source domain and uses its predictions on the target data as additional input features. A second model is then trained on the labelled target data, augmented with these features. For a fair comparison, the source model uses the same architecture as ADAPTOOD.

\textbf{Instance-Based Domain Adaptation.}
Further to the feature-based domain adaptation baseline seen above, we also use an instance-based domain adaptation alternative. Instead of researching common features, this approach focuses on reweighting training data in order to correct the difference between the source and target distributions. To implement this, we rely on the widely-used nearest neighbours weighting strategy, which reweights the source instances according to their number of neighbours in the target dataset~\cite{loog21}. During training, this reweighting involves multiplying the loss of each training instance by a positive weight. For a fair comparison, the estimator used to learn the task uses the same architecture as ADAPTOOD's model.

\textbf{Ablation using only the Hellinger Distance.}
To enhance our comprehension of which metric is more effective for estimating uncertainty through the calculation of distribution-level divergence, we conduct an ablation study employing solely the Hellinger distance. All other elements of ADAPTOOD are retained, thus ensuring that any observed performance improvements can be attributed exclusively to the Hellinger distance.

\textbf{Ablation using only the Mahalanobis Distance.}
Likewise, a separate ablation study is conducted using exclusively the Mahalanobis distance. Although our selected metrics are complementary, the sole use of the Mahalanobis distance highlights its distinct advantages over the Hellinger distance. All other elements of ADAPTOOD are retained, thus ensuring that any observed performance improvements can be attributed exclusively to the Mahalanobis distance metric.

\textbf{Ablation without Hyperparameter Tuning.}
To understand the impact of ADAPTOOD's additional components, this ablation removes its hyperparameter tuning. All other aspects remain unchanged, including the OOD adaptation mechanism and the LoRA module. This setup isolates the performance gains due to improved hyperparameter optimisation. To this end, the relevant hyperparameters are fixed and set to use the Adam optimiser and its default learning rate of $1\mathrm{e}{-3}$.

\textbf{Ablation without LoRA.}
To assess the effect of low-rank adaptation, we run a set of experiments by disabling it. These examine how it affects the adapted model in terms of size, memory footprint, and parameter count. All other aspects of the ADAPTOOD framework remain unchanged, including the OOD adaptation mechanism and the hyperparameter tuning module.
\section{Results} \label{results}
\begin{table*}[t]
    \centering
    \caption{Ablation study results, comparing variants of our approach with the full method}
    \label{tab:ablation_study_metrics}
    \begin{tabular}{l|cccccc}
        \toprule\toprule
        \textbf{Method} & \textbf{Accuracy} & \textbf{Precision} & \textbf{Recall} & \textbf{Sensitivity} & \textbf{Specificity} & \textbf{F1 Score}\\
        \midrule\midrule
        {} & \multicolumn{6}{c}{ECG MIT-BIH}\\
        \midrule
            Hellinger Distance & 0.930\,\scriptsize{$\pm$\,0.023} & 0.912\,\scriptsize{$\pm$\,0.011} & 0.832\,\scriptsize{$\pm$\,0.076} & 0.682\,\scriptsize{$\pm$\,0.157} & 0.981\,\scriptsize{$\pm$\,0.006} & 0.861\,\scriptsize{$\pm$\,0.058}\\
            Mahalanobis Distance & 0.951\,\scriptsize{$\pm$\,0.007} & 0.929\,\scriptsize{$\pm$\,0.012} & 0.894\,\scriptsize{$\pm$\,0.016} & 0.807\,\scriptsize{$\pm$\,0.030} & \textbf{0.981\,\scriptsize{$\pm$\,0.004}} & 0.910\,\scriptsize{$\pm$\,0.014}\\
            w/o HP Tuning & 0.945\,\scriptsize{$\pm$\,0.004} & 0.919\,\scriptsize{$\pm$\,0.026} & 0.887\,\scriptsize{$\pm$\,0.025} & 0.798\,\scriptsize{$\pm$\,0.061} & 0.976\,\scriptsize{$\pm$\,0.015} & 0.900\,\scriptsize{$\pm$\,0.009}\\
            w/o LoRA & 0.949\,\scriptsize{$\pm$\,0.004} & 0.915\,\scriptsize{$\pm$\,0.011} & \textbf{0.904\,\scriptsize{$\pm$\,0.009}} & \textbf{0.835\,\scriptsize{$\pm$\,0.018}} & 0.973\,\scriptsize{$\pm$\,0.006} & 0.910\,\scriptsize{$\pm$\,0.008}\\
            ADAPTOOD & \textbf{0.953\,\scriptsize{$\pm$\,0.002}} & \textbf{0.930\,\scriptsize{$\pm$\,0.008}} & 0.900\,\scriptsize{$\pm$\,0.005} & 0.820\,\scriptsize{$\pm$\,0.013} & 0.980\,\scriptsize{$\pm$\,0.005} & \textbf{0.914\,\scriptsize{$\pm$\,0.003}}\\
        \midrule\midrule
        {} & \multicolumn{6}{c}{ECG PTB-DB}\\
        \midrule
            Hellinger Distance & 0.825\,\scriptsize{$\pm$\,0.045} & 0.839\,\scriptsize{$\pm$\,0.024} & 0.719\,\scriptsize{$\pm$\,0.133} & 0.950\,\scriptsize{$\pm$\,0.058} & 0.488\,\scriptsize{$\pm$\,0.324} & 0.722\,\scriptsize{$\pm$\,0.129}\\
            Mahalanobis Distance & 0.875\,\scriptsize{$\pm$\,0.045} & 0.871\,\scriptsize{$\pm$\,0.034} & 0.800\,\scriptsize{$\pm$\,0.089} & \textbf{0.964\,\scriptsize{$\pm$\,0.007}} & 0.636\,\scriptsize{$\pm$\,0.185} & 0.819\,\scriptsize{$\pm$\,0.083}\\
            w/o HP Tuning & 0.882\,\scriptsize{$\pm$\,0.008} & 0.869\,\scriptsize{$\pm$\,0.009} & 0.822\,\scriptsize{$\pm$\,0.020} & 0.952\,\scriptsize{$\pm$\,0.014} & 0.691\,\scriptsize{$\pm$\,0.046} & 0.840\,\scriptsize{$\pm$\,0.014}\\
            w/o LoRA & 0.898\,\scriptsize{$\pm$\,0.008} & 0.885\,\scriptsize{$\pm$\,0.010} & 0.851\,\scriptsize{$\pm$\,0.017} & 0.954\,\scriptsize{$\pm$\,0.011} & 0.747\,\scriptsize{$\pm$\,0.037} & 0.865\,\scriptsize{$\pm$\,0.012}\\
            ADAPTOOD & \textbf{0.922\,\scriptsize{$\pm$\,0.013}} & \textbf{0.901\,\scriptsize{$\pm$\,0.019}} & \textbf{0.899\,\scriptsize{$\pm$\,0.015}} & 0.948\,\scriptsize{$\pm$\,0.014} & \textbf{0.852\,\scriptsize{$\pm$\,0.019}} & \textbf{0.900\,\scriptsize{$\pm$\,0.016}}\\
        \midrule\midrule
        {} & \multicolumn{6}{c}{ECG MIMICPERformTT}\\
        \midrule
            Hellinger Distance & 0.933\,\scriptsize{$\pm$\,0.028} & 0.938\,\scriptsize{$\pm$\,0.022} & 0.933\,\scriptsize{$\pm$\,0.028} & 0.980\,\scriptsize{$\pm$\,0.010} & 0.887\,\scriptsize{$\pm$\,0.060} & 0.933\,\scriptsize{$\pm$\,0.028}\\
            Mahalanobis Distance & 0.937\,\scriptsize{$\pm$\,0.005} & 0.942\,\scriptsize{$\pm$\,0.003} & 0.937\,\scriptsize{$\pm$\,0.005} & \textbf{0.990\,\scriptsize{$\pm$\,0.010}} & 0.883\,\scriptsize{$\pm$\,0.020} & 0.937\,\scriptsize{$\pm$\,0.005}\\
            w/o HP Tuning & 0.913\,\scriptsize{$\pm$\,0.020} & 0.923\,\scriptsize{$\pm$\,0.012} & 0.913\,\scriptsize{$\pm$\,0.020} & 0.983\,\scriptsize{$\pm$\,0.015} & 0.843\,\scriptsize{$\pm$\,0.050} & 0.913\,\scriptsize{$\pm$\,0.021}\\
            w/o LoRA & 0.918\,\scriptsize{$\pm$\,0.020} & 0.925\,\scriptsize{$\pm$\,0.015} & 0.918\,\scriptsize{$\pm$\,0.020} & 0.977\,\scriptsize{$\pm$\,0.005} & 0.860\,\scriptsize{$\pm$\,0.045} & 0.918\,\scriptsize{$\pm$\,0.021}\\
            ADAPTOOD & \textbf{0.942\,\scriptsize{$\pm$\,0.003}} & \textbf{0.944\,\scriptsize{$\pm$\,0.002}} & \textbf{0.942\,\scriptsize{$\pm$\,0.003}} & 0.980\,\scriptsize{$\pm$\,0.010} & \textbf{0.903\,\scriptsize{$\pm$\,0.015}} & \textbf{0.942\,\scriptsize{$\pm$\,0.003}}\\
        \midrule\midrule
        {} & \multicolumn{6}{c}{PPG MIMICPERformTT}\\
        \midrule
            Hellinger Distance & 0.958\,\scriptsize{$\pm$\,0.023} & 0.962\,\scriptsize{$\pm$\,0.020} & 0.958\,\scriptsize{$\pm$\,0.023} & 0.995\,\scriptsize{$\pm$\,0.005} & 0.920\,\scriptsize{$\pm$\,0.050} & 0.958\,\scriptsize{$\pm$\,0.023}\\
            Mahalanobis Distance & 0.965\,\scriptsize{$\pm$\,0.023} & 0.968\,\scriptsize{$\pm$\,0.020} & 0.965\,\scriptsize{$\pm$\,0.023} & 1.000\,\scriptsize{$\pm$\,0.000} & 0.930\,\scriptsize{$\pm$\,0.045} & 0.965\,\scriptsize{$\pm$\,0.023}\\
            w/o HP Tuning & 0.953\,\scriptsize{$\pm$\,0.025} & 0.958\,\scriptsize{$\pm$\,0.022} & 0.953\,\scriptsize{$\pm$\,0.025} & 1.000\,\scriptsize{$\pm$\,0.000} & 0.907\,\scriptsize{$\pm$\,0.050} & 0.953\,\scriptsize{$\pm$\,0.025}\\
            w/o LoRA & 0.952\,\scriptsize{$\pm$\,0.005} & 0.956\,\scriptsize{$\pm$\,0.005} & 0.952\,\scriptsize{$\pm$\,0.005} & 1.000\,\scriptsize{$\pm$\,0.000} & 0.903\,\scriptsize{$\pm$\,0.010} & 0.952\,\scriptsize{$\pm$\,0.005}\\
            ADAPTOOD & \textbf{0.975\,\scriptsize{$\pm$\,0.010}} & \textbf{0.976\,\scriptsize{$\pm$\,0.009}} & \textbf{0.975\,\scriptsize{$\pm$\,0.010}} & 0.995\,\scriptsize{$\pm$\,0.005} & \textbf{0.953\,\scriptsize{$\pm$\,0.025}} & \textbf{0.975\,\scriptsize{$\pm$\,0.010}}\\
        \midrule\midrule
        {} & \multicolumn{6}{c}{ECG CODEtest}\\
        \midrule
            Hellinger Distance & 0.914\,\scriptsize{$\pm$\,0.027} & 0.914\,\scriptsize{$\pm$\,0.023} & 0.799\,\scriptsize{$\pm$\,0.119} & 0.622\,\scriptsize{$\pm$\,0.252} & 0.977\,\scriptsize{$\pm$\,0.020} & 0.824\,\scriptsize{$\pm$\,0.100}\\
            Mahalanobis Distance & 0.912\,\scriptsize{$\pm$\,0.021} & 0.873\,\scriptsize{$\pm$\,0.035} & \textbf{0.824\,\scriptsize{$\pm$\,0.072}} & \textbf{0.683\,\scriptsize{$\pm$\,0.144}} & 0.965\,\scriptsize{$\pm$\,0.003} & \textbf{0.843\,\scriptsize{$\pm$\,0.057}}\\
            w/o HP Tuning & 0.906\,\scriptsize{$\pm$\,0.003} & 0.869\,\scriptsize{$\pm$\,0.024} & 0.753\,\scriptsize{$\pm$\,0.019} & 0.527\,\scriptsize{$\pm$\,0.039} & 0.978\,\scriptsize{$\pm$\,0.007} & 0.794\,\scriptsize{$\pm$\,0.020}\\
            w/o LoRA & 0.916\,\scriptsize{$\pm$\,0.024} & \textbf{0.927\,\scriptsize{$\pm$\,0.017}} & 0.792\,\scriptsize{$\pm$\,0.081} & 0.593\,\scriptsize{$\pm$\,0.173} & \textbf{0.990\,\scriptsize{$\pm$\,0.011}} & 0.831\,\scriptsize{$\pm$\,0.063}\\
            ADAPTOOD & \textbf{0.922\,\scriptsize{$\pm$\,0.018}} & 0.875\,\scriptsize{$\pm$\,0.053} & 0.807\,\scriptsize{$\pm$\,0.052} & 0.641\,\scriptsize{$\pm$\,0.093} & 0.974\,\scriptsize{$\pm$\,0.011} & 0.836\,\scriptsize{$\pm$\,0.053}\\
        \bottomrule\bottomrule
    \end{tabular}
\end{table*}

\subsection{Key Findings} \label{results:key_findings}
Under population, sensor, temporal and other shifts (discussed in Section~\ref{methods:problem_definition}) which occur for the datasets examined, ADAPTOOD achieves strong improvements by guiding adaptation based on OOD severity, indicating that explicitly accounting for OOD levels leads to stronger and more balanced generalisation.

In the MIT-BIH data, these shifts cause transfer learning to perform very closely to supervised learning in accuracy (0.910 vs 0.906) and precision (0.891 vs 0.914). Yet, our OOD severity-based mechanism improves performance across metrics, including in accuracy (0.953) and precision (0.930). While the sensitivity of these baselines remains near chance level (0.595 and 0.499), ADAPTOOD achieves 0.820 by tailoring its adaptation to OOD severity, leading to more generalisable classification. The feature-based domain adaptation baseline shows improvements, but ADAPTOOD still outperforms it in all cases, including in accuracy (0.953 vs 0.940), precision (0.930 vs 0.921), and recall (0.900 vs 0.861). Specificity is inflated for supervised learning compared to ADAPTOOD (0.991 vs 0.980), but this results from the baseline's tendency to underpredict the positive class, leading to unbalanced performance.

Similar results are observed in the PTB-DB data, where transfer learning and supervised learning perform nearly identically. The domain adaptation cases, which are more advanced by design, perform better, but ADAPTOOD still outperforms all. For instance, in accuracy (0.922) it achieves improvements of 11.9\% over the transfer learning (0.803), 10.4\% over the supervised learning (0.818), 4.7\% over the feature-based domain adaptation (0.875), and 4.9\% over the instance-based domain adaptation baselines. A similar trend is recorded in the recall (0.899 vs 0.797 vs 0.728 vs 0.836 vs 0.835), while in precision it achieves a 12.9\% increase compared to transfer learning (0.901 vs 0.772). These results show that ADAPTOOD's largest performance gains occur under more severe distributional shifts, validating the design of the severity-aware adaptation mechanism.

When, in addition to the above shifts, there is also a task and label shift, the OOD scenario becomes even more severe. This is the case for the ECG MIMICERformTT dataset, but ADAPTOOD continues to demonstrate superior generalisation. It achieves an accuracy of 0.942, significantly higher than both the transfer learning (0.882) and domain adaptation (0.867 and 0.842) baselines. This pattern extends to the F1 score, where ADAPTOOD reaches 0.942 compared to 0.881, 0.866, and 0.841 for the respective baselines, corresponding to an F1 improvement of 10.1\%. This demonstrates that the OOD severity-based mechanism maintains balanced improvements across accuracy and F1 score, even under compound distributional shifts.

To further test ADAPTOOD, we examine a different modality, specifically the PPG recordings of the MIMICPERform PPG dataset. This introduces a modality shift in addition to the shifts considered in previous scenarios, representing an edge case. In this case, transfer learning brings a marginal improvement of only 1\% in accuracy over the supervised baseline (0.917 vs 0.905), while the instance-based domain adaptation offers a better solution that results in 4\% higher accuracy (0.945 vs 0.905). Yet, ADAPTOOD outperforms all by reaching 0.975 in accuracy, while similar results are recorded in other metrics too, like the precision and the recall. Although fine-tuning is rarely the go-to solution when adapting models across modalities and ADAPTOOD is not designed for such cases, this experiment demonstrates its ability to handle extreme OOD conditions, achieving up to 7\% higher accuracy than the strongest baseline.

In the CODEtest dataset, which introduces a dimensionality shift by increasing the number of input channels, the results are also interesting. That's because it was originally recorded using 12-lead configurations of ECG biosignals, rather than single-lead like the previous wearable datasets. The transfer learning, supervised learning, feature-based domain adaptation and instance-based domain adaptation baselines reach values of 0.873, 0.861, 0.894, and 0.882 in accuracy, respectively, but ADAPTOOD outperforms all with a value of 0.922. Our solution's improvements also extend to other metrics, including in precision (0.875 vs less than 0.825 for the baselines) and F1 score (0.836 vs less than 0.789 for the baselines). Overall, ADAPTOOD consistently delivers gains in accuracy across diverse OOD scenarios, reinforcing that modelling OOD severity is a key driver of robust generalisation.

\subsection{Ablation Studies} \label{results:ablation_study}
While ADAPTOOD's uncertainty-guided transfer learning system serves as its core engine, both the choice of uncertainty metrics and the choice of its supporting components is imperative towards its success.

\textbf{Uncertainty Metric Comparisons.}
To understand which metric of Section~\ref{methods:uncertainty} is the most effective in capturing OOD-related uncertainty, we conduct an ablation study with distinct experiments: one utilising the Hellinger distance and another using the Mahalanobis distance. The results can be found in Table~\ref{tab:ablation_study_metrics}. While the Mahalanobis distance outperforms the Hellinger version, our approach that integrates both performs consistently better. For instance, in the PTB-DB data it reaches an accuracy of 0.922, compared to just 0.825 when using the Hellinger distance or 0.875 when using the Mahalanobis distance. Similarly, in the ECG MIMICPERformTT data, ADAPTOOD reaches a recall value of 0.942, while the Hellinger ablation has a lower value of 0.933. While the differences are not significant in the case of switching the metrics, combining them allows for more consistent results.

\textbf{Supporting Mechanism Comparisons.}
Further to the uncertainty-based ablations, we also conduct a study to understand the importance of its supporting components. We observe that both the hyperparameter (HP) tuning and the LoRA integration contribute to its performance gains. In ECG PTB-DB data, removing HP tuning lowers ADAPTOOD's accuracy by 4\% (0.922 vs 0.882), while this is also confirmed through other metrics like the precision (0.901 vs 0.869) and the recall (0.899 vs 0.822) that also get reduced. A similar drop is noted in ECG MIMICPERformTT, where the ablation without HP tuning shows a 3\% drop in accuracy (0.942 vs 0.913). However, the HP tuning module is not as significant in some other cases. For instance, its removal in the MIT-BIH data leads to less than 1\% drop in accuracy (0.953 vs 0.945). This shows that while selective layer unfreezing drives adaptation, HP tuning is also key when the pre-trained model is less suited to the OOD levels.

The significance of LoRA shows resembling behaviour: some of the ablations that do not use it suffer from negligible performance degradation, while for others it proves crucial. In the ECG MIMICPERformTT data, its removal decreases the accuracy and F1 score by 2.9\% (both 0.942 vs 0.913) and the specificity by as much as 6\% (0.903 vs 0.843). In contrast, in the ECG CODEtest data it results in just a 0.6\% decrease in accuracy (0.922 vs 0.916), while surprisingly improving precision (0.875 vs 0.927) compared to ADAPTOOD. Yet, these improvements are likely due to noise or dataset-specific variance. Therefore, using the full ADAPTOOD model, is recommended for consistent and robust adaptation under varying OOD severity scenarios.

\subsection{Model Efficiency} \label{results:efficiency}
ADAPTOOD is computationally efficient, as it leads to a substantial reduction in both total parameter count and memory footprint compared to the ablation without LoRA across all evaluated datasets.

Its computational efficiency is achieved through its uncertainty-based adaptation approach and the use of low-rank model updates that enable parameter-efficient adaptation. For instance, in the MIT-BIH dataset, ADAPTOOD's final model requires only 2,256,225 total parameters (8.61 MB) versus 7,647,781 parameters (29.17 MB) for the version without LoRA. Similar trends are observed for the PTB-DB (2,256,225 vs. 7,647,781 parameters; 8.61 MB vs. 29.17 MB), ECG MIMICPERformTT (21,392,737 vs. 26,784,293 parameters; 81.61 MB vs. 102.17 MB), PPG MIMICPERformTT (21,392,737 vs. 26,784,293 parameters; 81.61 MB vs. 102.17 MB), and CODEtest (14,806,369 vs. 20,197,925 parameters; 56.48 MB vs. 77.05 MB) data. Moreover, ADAPTOOD updates only the minimal set of parameters necessary, rather than fine-tuning the full model. Its design also emphasises adaptive updates: uncertainty-aware adjustments mean that minimal updates are made for high-confidence cases, while more extensive fine-tuning is performed for low-confidence cases. Thus, it is a lightweight yet powerful framework that leads to computational cost savings too.
\section{Conclusion} \label{conclusion}
ADAPTOOD introduces dynamic and uncertainty-aware fine-tuning of pre-trained models for time series data under out-of-distribution conditions. It directly addresses the complex and varied distribution shifts seen in real-world ECG biosignals, arising from differences in sensors, populations, domains, labels, and temporal contexts. By fine-tuning only when necessary it avoids overfitting and reduces unnecessary computation. Its adaptive strategy leverages uncertainty to guide selective layer training, applies LoRA to update relevant components, and uses hyperparameter optimisation to adjust settings effectively. 

The ADAPTOOD method outperforms alternatives across a broad range of OOD scenarios, showing consistent gains in robustness, calibration, and generalisation. Its ablations perform strongly as well, which is important as it shows that it can also be a reliable option in settings with varying scalability, complexity, or real-time deployment requirements. A key contributor to this performance is ADAPTOOD's ability to adapt the degree of fine-tuning based on the severity of distribution shifts, indicating the benefit of uncertainty-guided adaptation over static protocols. These results position ADAPTOOD as a promising and reliable approach for out-of-distribution model fine-tuning cases.

\section*{ACKNOWLEDGMENT}
This work is supported by Arm and by EPSRC grant EP/S023046/1 for the EPSRC Centre for Doctoral Training in Sensor Technologies and Applications.
\bibliographystyle{IEEEtran}
\bibliography{bibliography}

@article{gupta24,
title = {\href{https://doi.org/10.1016/j.heliyon.2024.e26787}{A Comprehensive Review on Efficient Artificial Intelligence Models for Classification of Abnormal Cardiac Rhythms using Electrocardiograms}},
author = {Utkarsh Gupta and Naveen Paluru and Deepankar Nankani and Kanchan Kulkarni and Navchetan Awasthi},
journal = {Heliyon},
volume = {10},
number = {5},
pages = {e26787},
year = {2024}
}

@article{ding25,
title = {\href{https://doi.org/10.1016/j.bios.2024.117073}{Advances in Deep Learning for Personalized ECG Diagnostics: A Systematic Review Addressing Inter-Patient Variability and Generalization Constraints}},
author = {Cheng Ding and Tianliang Yao and Chenwei Wu and Jianyuan Ni},
journal = {Biosensors and Bioelectronics},
volume = {271},
pages = {117073},
year = {2025}
}

@inproceedings{embc25,
title = {\href{https://doi.org/10.1109/EMBC58623.2025.11254027}{SALTS: Streamlined Adaptive Learning for Sensors Time Series}},
author = {Vavaroutas, Sotirios and Rizos, Georgios and Mascolo, Cecilia},
booktitle = {Proceedings of the 47th Annual International Conference of the IEEE Engineering in Medicine and Biology Society (EMBC)},
year = {2025}
}

@article{jingkang24,
title = {\href{https://doi.org/10.1007/s11263-024-02117-4}{Generalized Out-of-Distribution Detection: A Survey}},
author = {Yang, Jingkang and Zhou, Kaiyang and Li, Yixuan and Liu, Ziwei},
journal = {International Journal of Computer Vision},
pages = {5635–5662},
volume = {132},
number = {12},
numpages = {28},
year = {2024}
}

@article{rajendran24,
title = {\href{https://doi.org/10.1016/j.patter.2023.100913}{Learning Across Diverse Biomedical Data Modalities and Cohorts: Challenges and Opportunities for Innovation}},
author = {Suraj Rajendran and Weishen Pan and Mert R. Sabuncu and Yong Chen and Jiayu Zhou and Fei Wang},
journal = {Patterns},
volume = {5},
number = {2},
pages = {100913},
issn = {2666-3899},
year = {2024}
}

@article{xia22,
title = {\href{https://doi.org/10.1007/978-3-031-14771-5_25}{Benchmarking Uncertainty Quantification on Biosignal Classification Tasks Under Dataset Shift}},
author = {Xia, Tong and Han, Jing and Mascolo, Cecilia},
journal = {Springer Multimodal AI in Healthcare},
publisher = {Springer International Publishing},
pages = {347–359},
year = {2022}
}

@inproceedings{jun23,
title = {\href{https://doi.org/10.1145/3580305.3599576}{Trustworthy Transfer Learning: Transferability and Trustworthiness}},
author = {Wu, Jun and He, Jingrui},
booktitle = {Proceedings of the 29th ACM SIGKDD Conference on Knowledge Discovery and Data Mining},
publisher = {Association for Computing Machinery},
pages = {5829–5830},
numpages = {2},
year = {2023}
}

@article{roy19,
title = {\href{https://doi.org/10.1088/1741-2552/ab260c}{Deep Learning-Based Electroencephalography Analysis: A Systematic Review}},
author = {Roy, Yannick and Banville, Hubert and Albuquerque, Isabela and Gramfort, Alexandre and Falk, Tiago H and Faubert, Jocelyn},
journal = {Journal of Neural Engineering},
publisher = {IOP Publishing},
volume = {16},
number = {5},
pages = {051001},
year = {2019}
}

@article{lichao20,
title = {\href{https://doi.org/10.3389/fnhum.2020.00103}{Cross-Dataset Variability Problem in EEG Decoding with Deep Learning}},
author = {Xu, Lichao and Xu, Minpeng and Ke, Yufeng and An, Xingwei and Liu, Shuang and Ming, Dong},
journal = {Frontiers in Human Neuroscience},
publisher = {Frontiers Media SA},
volume = {14},
issn = {1662-5161},
year = {2020}
}

@article{kouw19,
title = {\href{https://arxiv.org/abs/1812.11806}{An Introduction to Domain Adaptation and Transfer Learning}}, 
author = {Wouter M. Kouw and Marco Loog},
journal = {arXiv},
volume = {1812.11806},
year = {2019}
}

@inproceedings{jiahao24,
title = {\href{https://proceedings.mlr.press/v235/ai24a.html}{Not All Distributional Ahifts are Equal: Fine-Grained Robust Conformal Inference}},
author = {Ai, Jiahao and Ren, Zhimei},
booktitle = {Proceedings of the 41st International Conference on Machine Learning (ICML)},
publisher = {PMLR},
volume = {235},
pages = {641--665},
year = {2024}
}

@article{lakshminarayanan17,
title = {\href{https://proceedings.neurips.cc/paper_files/paper/2017/file/9ef2ed4b7fd2c810847ffa5fa85bce38-Paper.pdf}{Simple and Scalable Predictive Uncertainty Estimation using Deep Ensembles}},
author = {Lakshminarayanan, Balaji and Pritzel, Alexander and Blundell, Charles},
journal = {Advances in Neural Information Processing Systems (NeurIPS)},
publisher = {Curran Associates, Inc.},
volume = {30},
year = {2017}
}

@article{ovadia19,
title = {\href{https://proceedings.neurips.cc/paper_files/paper/2019/file/8558cb408c1d76621371888657d2eb1d-Paper.pdf}{Can You Trust Your Model's Uncertainty? Evaluating Predictive Uncertainty Under Dataset Shift}},
author = {Ovadia, Yaniv and Fertig, Emily and Ren, Jie and Nado, Zachary and Sculley, D. and Nowozin, Sebastian and Dillon, Joshua and Lakshminarayanan, Balaji and Snoek, Jasper},
journal = {Advances in Neural Information Processing Systems (NeurIPS)},
publisher = {Curran Associates, Inc.},
volume = {32},
year = {2019}
}

@article{svensson25,
title = {\href{https://doi.org/10.1007/978-3-031-72381-0_11}{Temporal Evaluation of Uncertainty Quantification Under Distribution Shift}},
author = {Svensson, Emma and Friesacher, Hannah Rosa and Arany, Adam and Mervin, Lewis and Engkvist, Ola},
journal = {AI in Drug Discovery},
publisher = {Springer Nature Switzerland},
pages = {132--148},
year = {2025}
}

@article{gholizade25,
title = {\href{https://doi.org/10.1007/s13198-024-02684-2}{A Review of Recent Advances and Strategies in Transfer Learning}},
journal = {International Journal of System Assurance Engineering and Management},
author = {Gholizade, Masoume and Soltanizadeh, Hadi and Rahmanimanesh, Mohammad and Sana, Shib Sankar},
volume = {16},
issn = {0976-4348},
number = {3},
pages = {1123–1162},
publisher = {Springer Science and Business Media {LLC}},
year = {2025}
}

@article{hosna22,
title = {\href{https://doi.org/10.1186/s40537-022-00652-w}{Transfer Learning: A Friendly Introduction}},
author = {Hosna, Asmaul and Merry, Ethel and Gyalmo, Jigmey and Alom, Zulfikar and Aung, Zeyar and Azim, Mohammad Abdul},
journal = {Journal of Big Data},
publisher = {Springer Science and Business Media {LLC}},
volume = {9},
issn = {2196-1115},
number = {1},
year = {2022}
}

@article{althnian21,
title = {\href{https://doi.org/10.3390/app11020796}{Impact of Dataset Size on Classification Performance: An Empirical Evaluation in the Medical Domain}},
author = {Althnian, Alhanoof and AlSaeed, Duaa and Al-Baity, Heyam and Samha, Amani and Dris, Alanoud Bin and Alzakari, Najla and Abou Elwafa, Afnan and Kurdi, Heba},
journal = {Applied Sciences},
volume = {11},
year = {2021},
number = {2},
article-number = {796}
}

@article{bizzego21,
title = {\href{https://arxiv.org/abs/2110.13732}{Improving the Efficacy of Deep Learning Models for Heart Beat detection on Heterogeneous Datasets}},
author = {Andrea Bizzego and Giulio Gabrieli and Michelle Jin-Yee Neoh and Gianluca Esposito},
year = {2021}
}

@article{li25,
title = {\href{http://doi.org/10.1038/s41598-025-33057-9}{Research on Cross-Dataset Cardiac Signal Domain Generalization and Feature Interpretability}},
author = {Li, Runwen and Aierken, Yierpani and Xu, Yu and Liu, Jiang and Tang, Yongjiang},
year = {2025},
journal = {Scientific Reports},
volume = {16},
number = {1}
}

@article{huang24,
title = {\href{http://doi.org/10.1080/14796678.2024.2354082}{Generalization Challenges in Electrocardiogram Deep Learning: Insights from Dataset Characteristics and Attention Mechanism}},
author = {Huang, Zhaojing and MacLachlan, Sarisha and Yu, Leping and Herbozo Contreras, Luis Fernando and Truong, Nhan Duy and Ribeiro, Antonio Horta and Kavehei, Omid},
year = {2024},
journal = {Future Cardiology},
volume = {20},
number = {4},
pages = {209–220}
}

@article{bengio12,
title = {\href{https://arxiv.org/abs/1206.5538}{Representation Learning: A Review and New Perspectives}},
author = {Yoshua Bengio and Aaron C. Courville and Pascal Vincent},
journal = {IEEE Transactions on Pattern Analysis and Machine Intelligence},
volume = {35},
pages = {1798-1828},
year = {2012}
}

@article{wang23,
title = {\href{https://arxiv.org/abs/2209.07027}{Out-of-Distribution Representation Learning for Time Series Classification}},
author = {Wang Lu and Jindong Wang and Xinwei Sun and Yiqiang Chen and Xing Xie},
journal = {International Conference on Learning Representations (ICLR)},
year = {2023}
}

@article{mehari22,
title = {\href{https://doi.org/10.1016/j.compbiomed.2021.105114}{Self-Supervised Representation Learning from 12-Lead ECG Data}},
author = {Temesgen Mehari and Nils Strodthoff},
journal = {Computers in Biology and Medicine},
volume = {141},
pages = {105114},
year = {2022}
}

@article{trirat24,
title = {\href{https://arxiv.org/abs/2401.03717}{Universal Time-Series Representation Learning: A Survey}}, 
author = {Patara Trirat and Yooju Shin and Junhyeok Kang and Youngeun Nam and Jihye Na and Minyoung Bae and Joeun Kim and Byunghyun Kim and Jae-Gil Lee},
year = {2024}
}

@article{shu25,
title = {\href{https://arxiv.org/abs/2512.02180}{CLEF: Clinically-Guided Contrastive Learning for Electrocardiogram Foundation Models}},
author = {Yuxuan Shu and Peter H. Charlton and Fahim Kawsar and Jussi Hernesniemi and Mohammad Malekzadeh},
year = {2025}
}

@article{farahani21,
title = {\href{https://doi.org/10.1007/978-3-030-71704-9_65}{A Brief Review of Domain Adaptation}},
author = {Farahani, Abolfazl and Voghoei, Sahar and Rasheed, Khaled and Arabnia, Hamid R.},
journal = {Advances in Data Science and Information Engineering},
pages = {877--894},
year = {2021}
}

@inproceedings{chidlovskii16,
title = {\href{https://doi.org/10.1145/2939672.2939716}{Domain Adaptation in the Absence of Source Domain Data}},
author = {Chidlovskii, Boris and Clinchant, Stephane and Csurka, Gabriela},
booktitle = {Proceedings of the 22nd ACM SIGKDD International Conference on Knowledge Discovery and Data Mining (KDD)},
publisher = {Association for Computing Machinery},
pages = {451–460},
numpages = {10},
year = {2016}
}

@inproceedings{huan23,
title = {\href{https://proceedings.mlr.press/v202/he23b.html}{Domain Adaptation for Time Series under Feature and Label Shifts}},
author = {He, Huan and Queen, Owen and Koker, Teddy and Cuevas, Consuelo and Tsiligkaridis, Theodoros and Zitnik, Marinka},
booktitle = {Proceedings of the 40th International Conference on Machine Learning (ICML)},
volume = {202},
articleno = {518},
pages = {12746--12774},
numpages = {29},
year = {2023}
}

@article{yang20,
title = {\href{https://doi.org/10.1016/j.neucom.2020.07.061}{On Hyperparameter Optimization of Machine Learning Algorithms: Theory and Practice}},
author = {Li Yang and Abdallah Shami},
journal = {Neurocomputing},
volume = {415},
pages = {295-316},
issn = {0925-2312},
year = {2020}
}

@article{choi24,
title = {\href{https://arxiv.org/abs/2401.10220}{AutoFT: Learning an Objective for Robust Fine-Tuning}},
journal = {NeurIPS Workshop on Distribution Shifts},
author = {Caroline Choi and Yoonho Lee and Annie Chen and Allan Zhou and Aditi Raghunathan and Chelsea Finn},
year = {2024}
}

@article{tianyu24,
title = {\href{https://doi.org/10.1145/3631437}{RLoc: Towards Robust Indoor Localization by Quantifying Uncertainty}},
author = {Zhang, Tianyu and Zhang, Dongheng and Wang, Guanzhong and Li, Yadong and Hu, Yang and Sun, Qibin and Chen, Yan},
journal = {Proceedings of the ACM on Interactive, Mobile, Wearable and Ubiquitous Technologies (IMWUT)},
year = {2024},
publisher = {Association for Computing Machinery},
volume = {7},
number = {4},
articleno = {200},
numpages = {28}
}

@article{ziqi25,
title = {\href{https://doi.org/10.1145/3729488}{Incorporating Uncertainty in Predictive Models Using Mobile Sensing and Clinical Data: A Case Study on Persistent Post-surgical Pain}},
author = {Xu, Ziqi and Zhang, Jingwen and Haroutounian, Simon and Liu, Hanyang and Cao, Zihan and Messner, Gabrielle Rose and Alaverdyan, Harutyun B. and Ahuja, Saivee and Koshy, Rahul and Hanns, Joel and Frumkin, Madelyn and Rodebaugh, Thomas L. and Lu, Chenyang},
journal = {Proceedings of the ACM on Interactive, Mobile, Wearable and Ubiquitous Technologies (IMWUT)},
year = {2025},
publisher = {Association for Computing Machinery},
volume = {9},
number = {2},
articleno = {58},
numpages = {33}
}

@article{lopez25,
title = {\href{https://arxiv.org/abs/2505.02874}{Uncertainty Quantification for Machine Learning in Healthcare: A Survey}},
author = {L. Julián Lechuga López and Shaza Elsharief and Dhiyaa Al Jorf and Firas Darwish and Congbo Ma and Farah E. Shamout},
journal = {Conference on Health, Inference, and Learning (CHIL)},
year = {2025}
}

@inproceedings{huaxiu22,
title = {\href{https://arxiv.org/pdf/2211.14238}{Wild-Time: A Benchmark of In-the-Wild Distribution Shift over Time}},
author = {Yao, Huaxiu and Choi, Caroline and Cao, Bochuan and Lee, Yoonho and Koh, Pang Wei W and Finn, Chelsea},
booktitle = {Advances in Neural Information Processing Systems (NeurIPS)},
pages = {10309--10324},
publisher = {Curran Associates, Inc.},
volume = {35},
year = {2022}
}

@inproceedings{yuzhe23,
title = {\href{https://proceedings.mlr.press/v202/yang23s.html}{Change is Hard: A Closer Look at Subpopulation Shift}},
author = {Yang, Yuzhe and Zhang, Haoran and Katabi, Dina and Ghassemi, Marzyeh},
booktitle = {Proceedings of the 40th International Conference on Machine Learning (ICML)},
publisher = {PMLR},
pages = {39584--39622},
volume = {202},
year = {2023}
}

@article{simons20,
title = {\href{https://doi.org/10.1109/SMC42975.2020.9282912}{Impact of Physiological Sensor Variance on Machine Learning Algorithms}},
author = {Simons, Ama and Doyle, Thomas and Musson, David and Reilly, James},
journal = {IEEE International Conference on Systems, Man, and Cybernetics (SMC)},
pages = {241-247},
year = {2020}
}

@article{spathis22,
title = {\href{https://doi.org/10.1038/s41746-022-00719-1}{Longitudinal Cardio-Respiratory Fitness Prediction through Wearables in Free-Living Environments}},
author = {Spathis, Dimitris and Perez-Pozuelo, Ignacio and Gonzales, Tomas I. and Wu, Yu and Brage, Soren and Wareham, Nicholas and Mascolo, Cecilia},
journal = {npj Digital Medicine},
publisher = {Springer Science and Business Media {LLC}},
volume = {5},
issn = {2398-6352},
number = {1},
year = {2022}
}

@inproceedings{mouxiang24,
title = {\href{https://doi.org/10.1145/3637528.3671926}{Calibration of Time-Series Forecasting: Detecting and Adapting Context-Driven Distribution Shift}},
author = {Chen, Mouxiang and Shen, Lefei and Fu, Han and Li, Zhuo and Sun, Jianling and Liu, Chenghao},
booktitle = {Proceedings of the 30th ACM SIGKDD Conference on Knowledge Discovery and Data Mining (KDD)},
publisher = {Association for Computing Machinery},
pages = {341–352},
numpages = {12},
year = {2024}
}

@article{maesschalck20,
title = {\href{https://doi.org/10.1016/S0169-7439(99)00047-7}{The Mahalanobis Distance}},
journal = {Chemometrics and Intelligent Laboratory Systems},
author = {R. {De Maesschalck} and D. Jouan-Rimbaud and D.L. Massart},
volume = {50},
number = {1},
pages = {1-18},
issn = {0169-7439},
year = {2000}
}

@article{venkataramanan23,
title = {\href{https://doi.org/10.1109/ICCVW60793.2023.00483}{Gaussian Latent Representations for Uncertainty Estimation using Mahalanobis Distance in Deep Classifiers}},
author = {Venkataramanan, Aishwarya and Benbihi, Assia and Laviale, Martin and Pradalier, Cedric},
journal = {IEEE/CVF International Conference on Computer Vision Workshops (ICCVW)},
pages = {4490-4499},
publisher = {IEEE Computer Society},
year = {2023}
}

@article{govindaraj22,
title = {\href{https://doi.org/10.2139/ssrn.4035007}{The Hellinger Distance and its Applications to Hypothesis Testing and Model Uncertainty}},
author = {Govindaraj, Suresh and Tejas, Tavish},
journal = {SSRN Electronic Journal},
issn = {1556-5068},
publisher = {Elsevier BV},
year = {2022}
}

@article{zheng21,
title = {\href{https://doi.org/10.1016/j.cnsns.2021.105720}{Quantifying Model Uncertainty for the Observed Non-Gaussian Data by the Hellinger Distance}},
author = {Yayun Zheng and Fang Yang and Jinqiao Duan and Jürgen Kurths},
journal = {Communications in Nonlinear Science and Numerical Simulation},
volume = {96},
pages = {105720},
issn = {1007-5704},
year = {2021}
}

@article{danielsson80,
title = {\href{https://doi.org/10.1016/0146-664X(80)90054-4}{Euclidean Distance Mapping}},
author = {Per-Erik Danielsson},
journal = {Computer Graphics and Image Processing},
volume = {14},
number = {3},
pages = {227-248},
issn = {0146-664X},
year = {1980}
}

@inproceedings{gal16,
title = {\href{https://proceedings.mlr.press/v48/gal16.html}{Dropout as a Bayesian Approximation: Representing Model Uncertainty in Deep Learning}},
author = {Gal, Yarin and Ghahramani, Zoubin},
booktitle = {Proceedings of The 33rd International Conference on Machine Learning},
pages = {1050--1059},
volume = {48},
year = {2016}
}

@article{kiranyaz21,
title = {\href{https://doi.org/10.1016/j.ymssp.2020.107398}{1D Convolutional Neural Networks and Applications: A Survey}},
author = {Serkan Kiranyaz and Onur Avci and Osama Abdeljaber and Turker Ince and Moncef Gabbouj and Daniel J. Inman},
journal = {Mechanical Systems and Signal Processing},
volume = {151},
pages = {107398},
year = {2021}
}

@article{hu22,
title = {\href{https://arxiv.org/abs/2106.09685}{LoRA: Low-Rank Adaptation of Large Language Models}},
author = {Edward J Hu and Yelong Shen and Phillip Wallis and Zeyuan Allen-Zhu and Yuanzhi Li and Shean Wang and Lu Wang and Weizhu Chen},
journal = {International Conference on Learning Representations (ICLR)},
year = {2022}
}

@article{snoek12,
title = {\href{https://arxiv.org/abs/1206.2944}{Practical Bayesian Optimization of Machine Learning Algorithms}},
author = {Jasper Snoek and Hugo Larochelle and Ryan P. Adams},
year = {2012}
}

@article{roy23,
title = {\href{http://doi.org/10.1007/s11334-023-00540-3}{Hyperparameter Optimization for Deep Neural Network Models: A Comprehensive Study on Methods and Techniques}},
author = {Roy, Sunita and Mehera, Ranjan and Pal, Rajat Kumar and Bandyopadhyay, Samir Kumar},
journal = {Innovations in Systems and Software Engineering},
publisher = {Springer},
year = {2023}
}

@article{clifford17,
title = {\href{https://doi.org/10.22489/CinC.2017.065-469}{AF Classification from a Short Single Lead ECG Recording: the Physionet Computing in Cardiology Challenge 2017}},
journal = {Computing in Cardiology Conference (CinC)},
author = {Clifford, Gari and Liu, Chengyu and Moody, Benjamin and Lehman, Li-wei and Silva, Ikaro and Li, Qiao and Johnson, Alistair and Mark, Roger},
issn = {2325-887X},
year = {2017}
}

@article{clifford17dataset,
title = {\href{https://www.kaggle.com/datasets/luigisaetta/physionet2017ecg}{Physionet 2017 {ECG}}},
author = {Clifford, Gari and Liu, Chengyu and Moody, Benjamin and Lehman, Li-wei and Silva, Ikaro and Li, Qiao and Johnson, Alistair and Mark, Roger},
journal = {Kaggle},
year = {2017}
}

@article{moody01,
title = {\href{https://doi.org/10.1109/51.932724}{The Impact of the MIT-BIH Arrhythmia Database}},
author = {Moody, G. B. and Mark, R. G.},
journal = {IEEE Engineering in Medicine and Biology},
volume = {20},
number = {3},
pages = {45-50},
year = {2001}
}

@article{moody01dataset,
title = {\href{https://www.physionet.org/content/mitdb/}{MIT-BIH Arrhythmia Database}},
journal = {PhysioNet},
author = {Moody, George and Mark, Roger},
year = {2005}
}

@article{bousseljot95,
title = {\href{https://doi.org/10.1515/bmte.1995.40.s1.317}{Nutzung der {EKG-Signaldatenbank} Cardiodat der {PTB} {\"u}ber das Internet}},
author = {Ralf Bousseljot and Dieter Kreiseler and Allard Schnabel},
journal = {De Gruyter Brill},
volume = {40},
pages = {317--318},
year = {1995}
}

@article{bousseljot95dataset,
title = {\href{https://www.physionet.org/content/ptbdb/}{PTB Diagnostic ECG Database}},
author = {Ralf Bousseljot},
journal = {PhysioNet},
year = {2004}
}

@article{moody20,
title = {\href{https://doi.org/10.13026/C2607M}{MIMIC-III Waveform Database}},
author = {Moody, Benjamin and Moody, George and Villarroel, Mauricio and Clifford, Gari and Silva, Ikaro},
journal = {PhysioNet},
year = {2020}
}

@article{charlton22,
title = {\href{https://doi.org/10.1088/1361-6579/ac826d}{Detecting Beats in the Photoplethysmogram: Benchmarking Open-Source Algorithms}},
author = {Charlton, Peter H and Kotzen, Kevin and Mejía-Mejía, Elisa and Aston, Philip J and Budidha, Karthik and Mant, Jonathan and Pettit, Callum and Behar, Joachim A and Kyriacou, Panicos A},
journal = {Physiological Measurement},
publisher = {IOP Publishing},
volume = {43},
issn = {1361-6579},
number = {8},
pages = {085007},
year = {2022}
}

@article{charlton22dataset,
title = {\href{https://doi.org/10.5281/ZENODO.6807402}{MIMICPERform Datasets}},
author = {Charlton, Peter H},
journal = {Zenodo},
year = {2022}
}

@article{ribeiro20,
title = {\href{https://doi.org/10.1038/s41467-020-15432-4}{Automatic Diagnosis of the 12-Lead {{ECG}} Using a Deep Neural Network}},
author = {Ribeiro, Ant{\^o}nio H. and Ribeiro, Manoel Horta and Paix{\~a}o, Gabriela M. M. and Oliveira, Derick M. and Gomes, Paulo R. and Canazart, J{\'e}ssica A. and Ferreira, Milton P. S. and Andersson, Carl R. and Macfarlane, Peter W. and Meira Jr., Wagner and Sch{\"o}n, Thomas B. and Ribeiro, Antonio Luiz P.},
journal = {Nature Communications},
year = {2020},
volume = {11},
pages = {1760},
number = {1}
}

@article{ribeiro20dataset,
title = {\href{https://doi.org/10.5281/ZENODO.3625006}{CODE-test: An Annotated 12-Lead ECG Dataset}},
author = {Ribeiro, Antonio H and Ribeiro, Manoel Horta and Paixão, Gabriela M. and Oliveira, Derick M. and Gomes, Paulo R. and Canazart, Jéssica A. and Ferreira, Milton P. and Andersson, Carl R. and Macfarlane, Peter W. and Meira Jr., Wagner and Sch\"{o}n, Thomas B. and Ribeiro, Antonio Luiz P.},
journal = {Zenodo},
year = {2020}
}

@article{fazeli18dataset,
title = {\href{https://www.kaggle.com/datasets/shayanfazeli/heartbeat}{ECG Heartbeat Categorization Dataset}},
journal = {Kaggle},
author = {Fazeli, Shayan},
year = {2018}
}

@inproceedings{daume07,
title = {\href{https://aclanthology.org/P07-1033/}{Frustratingly Easy Domain Adaptation}},
author = {Daum{\'e} III, Hal},
booktitle = {Proceedings of the 45th Annual Meeting of the Association of Computational Linguistics},
pages = {256--263},
year = {2007}
}

@inproceedings{loog21,
title = {\href{https://doi.org/10.1109/MLSP.2012.6349714}{Nearest Neighbor-Based Importance Weighting}},
author = {Loog, Marco},
booktitle = {2012 IEEE International Workshop on Machine Learning for Signal Processing}, 
year = {2012}
}

@article{kingma14,
title = {\href{https://arxiv.org/abs/1412.6980}{Adam: A Method for Stochastic Optimization}},
author = {Diederik P. Kingma and Jimmy Ba},
journal = {International Conference on Learning Representations (ICLR)},
year = {2015}
}

@article{mathelin25,
title = {\href{http://jmlr.org/papers/v26/23-1359.html}{Deep Out-of-Distribution Uncertainty Quantification via Weight Entropy Maximization}},
author = {Antoine de Mathelin and Fran{\c{c}}ois Deheeger and Mathilde Mougeot and Nicolas Vayatis},
journal = {Journal of Machine Learning Research},
year = {2025},
volume = {26},
number = {4}
}
\end{document}